\documentclass{article}
\PassOptionsToPackage{numbers, compress}{natbib}
 \usepackage[final]{nips_2017}

\usepackage[textsize=tiny]{todonotes}

\usepackage[utf8]{inputenc} 
\usepackage[T1]{fontenc}    
\usepackage{hyperref}       
\usepackage{url}            
\usepackage{booktabs}       
\usepackage{amsfonts}       
\usepackage{nicefrac}       
\usepackage{microtype}      
\usepackage{multirow}

\title{Modeling Preemptive Behaviors for Uncommon Hazardous Situations From Demonstrations}

\author{
  Priyam Parashar\thanks{http://acsweb.ucsd.edu/~pparasha/} \\
  Contextual Robotics Institute\\
  University of California, San Diego\\
  La Jolla, CA 92093 \\
  \texttt{pparashar@ucsd.edu} \\
  \And
  Akansel Cosgun, Alireza Nakhaei and Kikuo Fujimura\\
  Honda Research Institute, USA \\
  Mountain View, CA 94043\\
  \texttt{acosgun@honda-ri.com, anakhaei@honda-ri.com, kfujimura@honda-ri.com} \\
}

\begin{document}

\maketitle

\begin{abstract}
This paper presents a learning from demonstration approach to programming safe, autonomous 
behaviors for uncommon driving scenarios. Simulation is used to re-create a targeted driving 
situation, one containing a road-side hazard creating significant occlusion in an 
urban neighborhood, and collect optimal driving behaviors from 24 users.
Paper employs a key-frame based approach combined with an algorithm to linearly combine models 
in order to extend the behavior to novel variations of the target situation.
This approach is theoretically agnostic to the kind of LfD framework used for modeling data
and our results suggest it generalizes well to variations containing additional 
number of hazards occurring in sequence. The linear combination algorithm is informed by 
analysis of driving data, which also suggests that decision making algorithms need to
consider a trade-off between road-rules and immediate rewards to tackle some complex cases.
\end{abstract}

\section{Introduction}

\begin{figure}[h]
  \centering
  \includegraphics[width=\textwidth]{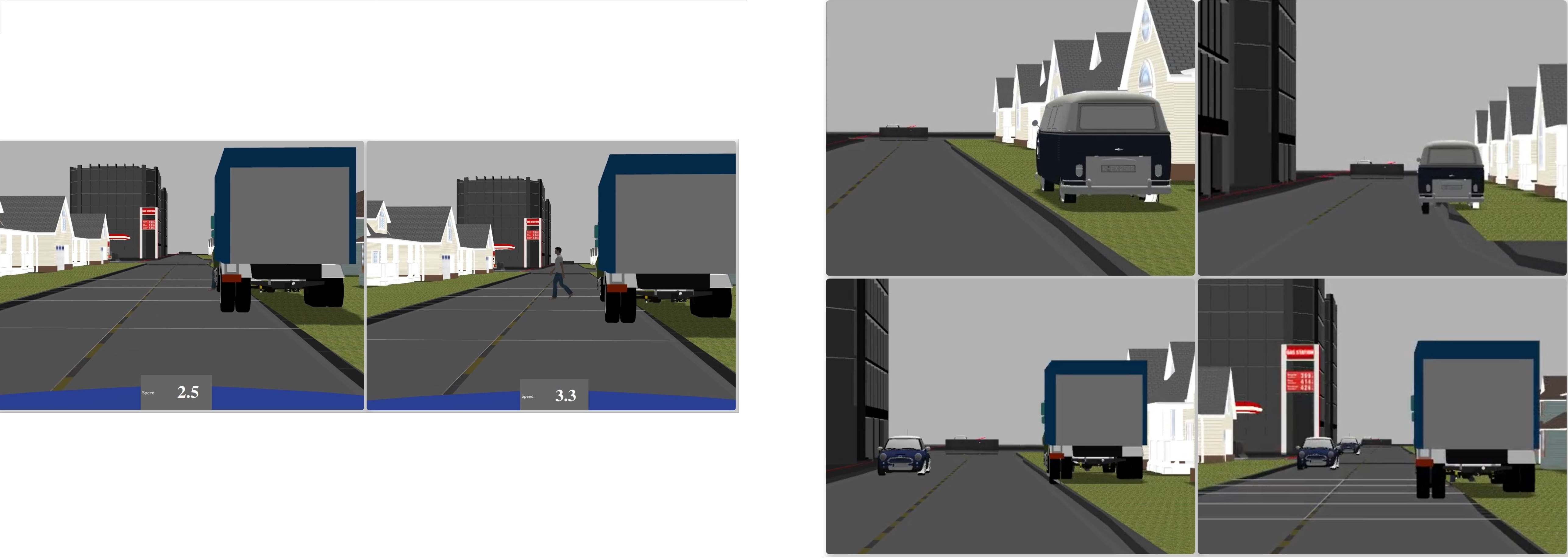}
  \caption{Depiction of Hazard and Variation of Size and Position in Simulation}
  \label{fig:hazards}
\end{figure}

There have been significant improvements in the field of autonomous driving 
\cite{bojarski2016end,chen2015deepdriving,zhang2016learning,sadigh2016planning,wei2013towards,ross2011reduction}, however we do not currently see such vehicles on our 
roads. The technical reason is that Auto Vehicles (AV) are expected to demonstrate 
safety records superior to humans. Driving on urban roads in 
common and predictable situations can  be considered a solved problem, 
however the real challenge of AV is handling unexpected situations while 
maintaining safety\cite{mitsch2013provably}.
In this work, we focus on one such situation: when there is a risk of a previously 
unobserved  pedestrian or object suddenly appearing from an occluded area (see figure 
\ref{fig:hazards}). Rather than waiting for the hazard to emerge, AVs can potentially 
take preemptive actions to reduce risk of an accident once a hazard is sensed, 
for instance by steering farther away or reducing speed\cite{mckenna1999hazard}. This paper studies how such
behaviors could be learned from human demonstrations.

End-to-end models \cite{bojarski2016end,chen2015deepdriving} are data-driven and generally suffer from disproportionate 
quantities of various corner-cases in datasets, being incomplete in terms of 
safety guarantees. Another viable approach is to recreate the corner cases in 
simulation and use approaches like reinforcement learning (RL), inverse RL or 
learning from demonstration to model the behavior. RL however either requires
an accurate model of the environment\cite{zhang2016learning} or large 
amount of exploration\cite{isele2017navigating} before figuring out how to 
avoid fatal scenarios, inverse 
RL on the other hand assumes similar underlying rewards for all demonstrations 
thus requiring large amount of perfect data to converge\cite{ziebart2008maximum}.

We use Learning from Demonstrations (LfD)\cite{argall2009survey,atkeson1997robot,akgun2012keyframe} in this work because it inherently 
captures the human knowledge of reacting to obvious as well as emerging 
hazardous situations without placing any limiting assumptions on the behavior. 
We conduct our experiments in simulation, however our framework is also easily 
applicable to a real vehicle, if the data is sourced from such demonstrations.

This is a hypothesis driven exploratory study. We present our results on how
the independent features of hazard affect behavior of drivers and also show
algorithm generated constraints from our trained models which learns variance of behavior
over the recorded behaviors. We have also added a generalization algorithm 
to the constraint generator, agnostic of LfD model used, which linearly combines 
models depending upon the environment to generate constraints for out-of-training-distribution (OOTD)
cases. Our results suggest that this approach captures enough 
details to generalize the solution for generating constraints given 
novel scenarios with multiple hazards in occurring in sequence with some or no overlap.

\section{Approach}
We started this work with a hypothesis that the size and closeness 
of the hazard have a significant correlation with the evasive 
behavior that it elicits in a human driver. By evasive behavior we mean, 
categorically, the extent of deviation from ``normal" driving trajectory, 
how early on this behavior is triggered ($D_{thresh}$) and the change in speed. To 
test this hypothesis, we implemented a driving simulator using Gazebo 
\footnote{http://gazebosim.org} as can be seen in figure \ref{fig:hazards}.

\subsection{System Architecture}

\begin{figure}
  \centering
  \includegraphics[width=0.6\textwidth]{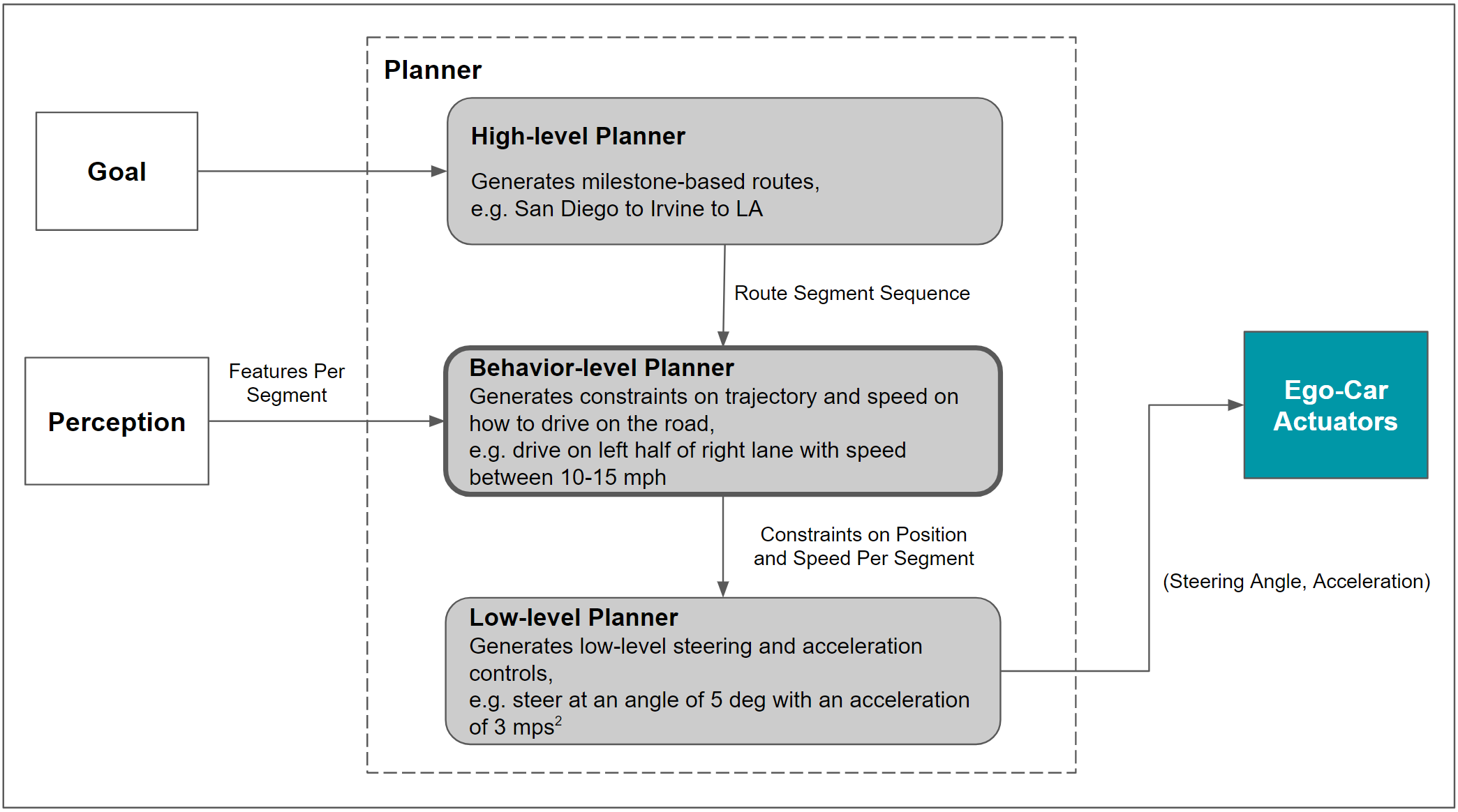}
  \caption{Schematic Diagram of System Architecture}
  \small
  This figure shows the three-tiered planning system as well as the data-flow between different 
  modules relevant to this paper. The mid-level planner is highlighted to emphasize algorithm placement.
  \label{fig:architecture}
\end{figure}

The system follows a funneled plan generation paradigm. The top layer 
provides a milestone-based plan using road networks to travel from start to
goal position, which we will call the \textit{route}. The middle layer generates speed, 
acceleration, trajectory, etc. constraints for the road-segments included in this route.
The bottom-most layer consists of the actual low-level controllers of the autonomous car and 
generates optimized trajectories using the constraints from mid-level planner. 
Figure \ref{fig:architecture} is a schematic illustration of this architecture and 
how it interacts with other relevant modules.

Our component sits in the middle layer and generates trajectory and speed
constraints for the current segment of the route for the particular case
of hazard occlusion. It depends on the perception 
module to provide it with the required state information (as described in section 
\ref{sec:key-frames}).
Middle layer consists of many such modules which generate constraints
and then the \textit{constraint aggregator} combines the ones it deems relevant based on its
scene understanding.

\subsection{Study and Data Collection}

Our driver population consisted of 24 people from Honda Research Institute, consisting 
of 3 female and 21 male drivers. More than 50\% of the population had a driving
experience of 10 years or plus and around 90\% of the population drove an 
average of 0 to 2 hours per week. The users were provided with one pilot run 
to acclimate themselves with the sensitivity of the wheel and feel of the simulator 
before having them drive the controlled test cases. Our recorded feature set 
consisted of global measures namely: Location of the car, location of the hazard, 
heading of the car, speed of the car, size of the hazard, nature of the road: 
bidirectional or unidirectional, and road lane limits. Users were also asked to 
fill out a survey to note their driving experience statistics and general demographic information.

We used a Logitech Driving Force G29 Racing wheel and paddle setup, to interface with 
the gazebo environment.
The interface for the study was completely based on Gazebo for visualization 
with ROS handling the back-end processing and communication. 
We wanted the drivers to have an idea that there is a non-zero probability of human beings 
appearing on the street, so that they drive with a safety-primed perspective to
account for the possibility of pedestrians behind the hazard. 
This was done indirectly by lining 
pathways with moving pedestrians. During the pilot run, we also added actual pedestrians 
crossing the streets from the pathway and also from behind vans to directly prime them for 
such a possibility. For actual data collection cases a pedestrian was programmed to walk out
from behind the hazard with a probability of 30\%.

For the controlled cases we manipulated the following independent categorical variables: (a) Size of the 
vehicle: Moderate (Van) or Large (truck), (b) Closeness of vehicle to the driving lane: Close ($\sim$ 10\%
of the vehicle parked on the road) or Far (vehicle clearly parked on the curb), (c) Direction of Traffic: 
No traffic (unidirectional road) or Opposing traffic (bidirectional road).
This gives us 8 unique scenarios which the user was required to drive through in the study. 
Figure \ref{fig:hazards} shows how the first two aspects were varied in the Gazebo world. Our 
dependent variables were: the choice of sub-lane on the road and the speed of the car.

In order to find the significance of effect that size and position of hazard 
had on driver's behavior we used paired t-test and Wilcoxon's test on the 
non-collision runs for each user. For the data measures to be of equal cardinality 
irrelevant of speed and sampling frequency, we binned speed and \textit{sub-lane} 
values over every 0.5 meters and averaged them. Sub-lanes are lanes were further 
categorized into 0.2 m wide strips. We only consider the data after $D_{thresh}$ 
has been crossed by the ego-car. We did this separately for each dependent 
variable. We only compared observations under similar traffic conditions.

\section{Training and Generating Constraints}

We adopted Key-frame based Learning from Demonstration (KLfD) from \citet{akgun2012keyframe} 
for modeling our training trajectories. It works by identifying key-frames 
in trajectories by repeatedly splining them based on base-set of knots, comparing
interpolated trajectory with the original and 
adding points of maximum error to the set until this comparison error is less than 
some threshold. The base knot set consists of the start and end-points of the 
user demonstrated trajectory and the final set is then termed as \textit{key-frames}. 
These key-frames are then further time aligned using Dynamic Time Warping 
\cite{berndt1994using} and clustered to give mean key-frames which can be used to 
extrapolate the behavior trajectory at run-time.

We used key-frames as an indirect measure of how far the ego-agent has progressed
in its behavior. The more distance ego-car has traveled with respect to the hazard,
the further it is in its behavior execution. We were able to use such a simplistic
metric only because we are only considering variations of given target case. However,
if an oracle exists which can provide our system with the closest key-frame to current 
state, this method will work irrelevant of level of data abstraction.

\subsection{Training}
\label{sec:key-frames}

Before starting key-frame extraction, the recorded features 
are first transformed to an intermediate, hazard-centric representation
(see sub-figure (a), (b) of figure \ref{fig:results}). We are following ROS conventions here, which means 
forward of ego-car (trajectory axis) is positive $y$ and left of ego-car 
(sub-lane axis) is positive $x$. After transformation the feature pile consists of:  
 Sub-lane ego-car is on, Lateral and perpendicular distance of ego-car from the hazard, 
 $V_{x}, V_{y}$ of the ego-car, Size of hazard, and Traffic condition.

Once transformed into this feature-set, we use only the non-accidental 
``good" (with points within road limits) trajectories and we train one model per unique scenario resulting in 
8 different models.
We follow the same steps as KLfD approach, except we use cubic splines (trade-off
between smoothness and ability to linearly combine many knots) and save 
the mean as well as the variance of clustered key-frames since we are 
primarily interested in the safety boundaries. This ordered tuple of 
$\big(\mu_{K_{x}}, \sigma_{K_{x}}\big)$ is saved as the scenario model.

\subsection{Constraint Generation}

The final framework takes the following features as input: Current sub-lane of the car,
position of the hazard, current heading of the car, current speed of the car, size of 
hazard, and the type of road.
The output of the framework is max and min limits on the sub-lanes as well as 
vehicle speed for a time horizon of 5 seconds. For evaluation purposes, we are not
classifying the cases but only passing randomized start values to each case model.

The system first calculates distance horizon by using current speed along with
time horizon of 5 seconds in ego-centric coordinate system. This distance is used 
to create a grid in hazard-centric coordinate system, where key-frame variances
are splined to create envelopes with respect to the hazard's location. Finally 
these envelopes are converted back to ego-car's coordinate system and passed
to constraint aggregator.

\subsubsection{Only One Hazard}

This is the simplest case, where we load the right model and use cubic splines
to extrapolate the envelope based on one standard deviation in both directions, 
accounting for $\sim$ 70\% of the variation.

\subsubsection{Multiple Hazards}

We treat each hazard as a new isolated one once we hit the $D_{thresh}$ for it. If 
the hazards are at enough distance from each other such that the latter's $D_{thresh}$
and formers active behavior time do not collide, then the agent just treats them 
as several one-hazard case and applies treatment from previous section. However, if 
hazards are close enough that the active behavior time and distance threshold overlap then
the agents treats this as two hazards together. The algorithm in this case, uses an
adapted $D_{thresh}$ for next hazard. This provides us with a mixed key-frame set 
consisting of previous hazard's key-frames lying before adapted $D_{thresh}$ and the 
next hazards key-frames lying after $D_{thresh}$. We use piecewise cubic splines
to smoothly combine these knots from different key-frame sets. Cubic splines are heavily favored 
in field like 
computer graphics \cite{fritsch1980monotone,brodlie1991preserving} because they are 
twice differentiable and produce smoother curves as compared to higher degree splines,
which is desirable in trajectory generation.

For hazard without any overlap we use end of the first hazard's bounding box as the
adapted $D_{thresh}$ and for hazards with overlap in bounding boxes we use the
average of their centers as the adapted $D_{thresh}$ (refer to section \ref{sec:resultsanddiscussion} for why we follow this methodology for combination).

\section{Results and Discussion}
\label{sec:resultsanddiscussion}

\begin{figure}[h]
  \centering
  \includegraphics[width=\textwidth]{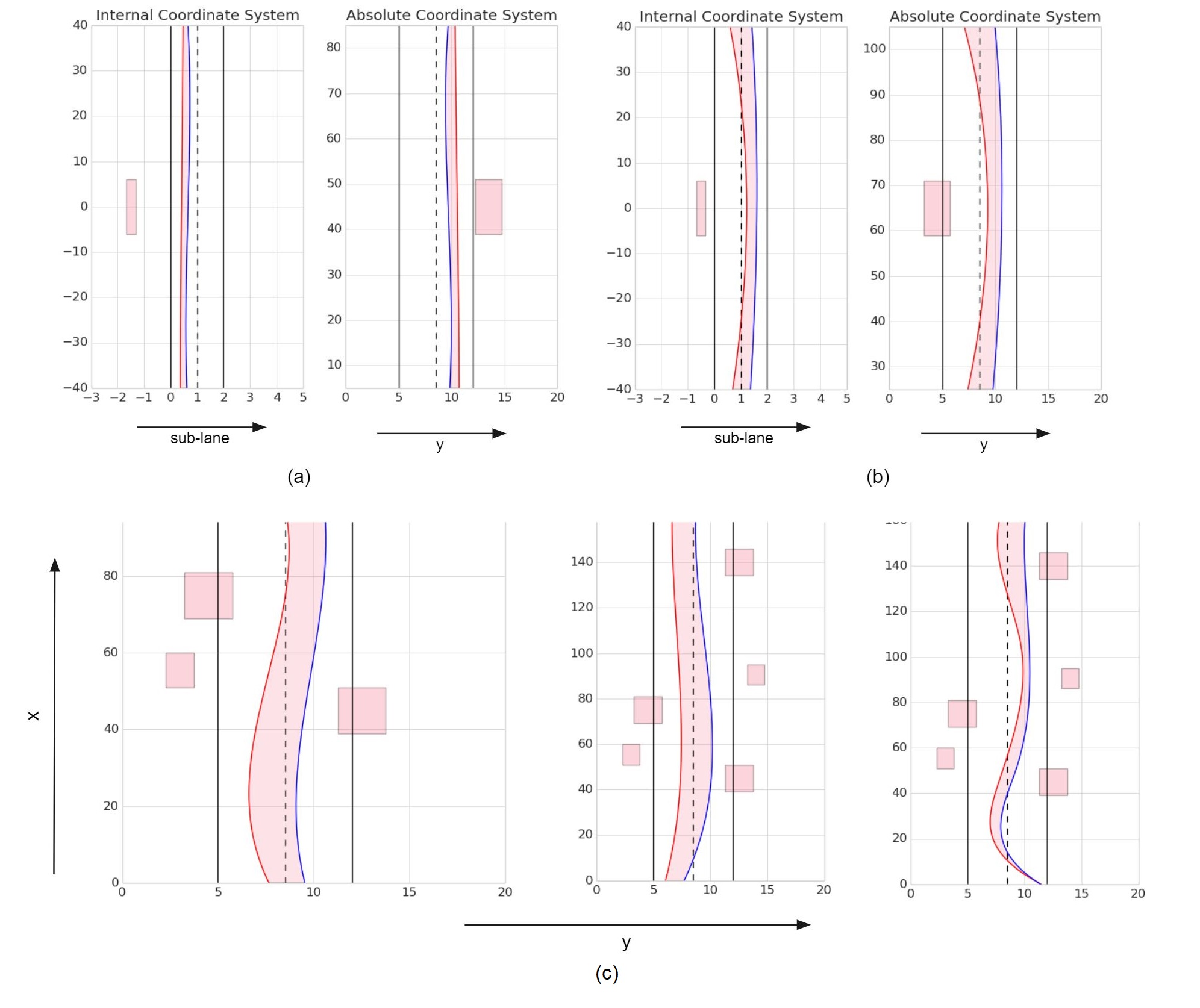}
  \caption{Generated Constraints for In-Distribution and OOTD Cases}
  \label{fig:results}
  \small
  (a) Constraints for large hazard in far position under bidirectional condition, (b) Constraints for
  large hazard in near position under unidirectional condition, (c) Constraints for OOTD cases arranged
  in order of complexity from left to right. They feature additional number of hazards and the last one
  also incorporates the bidirectional constraint of the road.
\end{figure}

\begin{table}[b]
  \centering
  \caption{Table showing $D_{thresh}$ and point of Maximum Curvature for various cases}
  \label{table:insight}
  \begin{tabular}{cccc}
    \hline
    Traffic & \multicolumn{1}{p{2cm}}{\centering Independent \\ Variable} & \multicolumn{1}{p{2cm}}{\centering $D_{thresh}$ \\ in m} & \multicolumn{1}{p{5cm}}{\centering Point of Maximum Curvature \\ (distance from hazard in m)}\\
    \hline
    \multirow{4}{*}{Uni} & \multicolumn{1}{c}{Small} & \multicolumn{1}{c}{36.5} & \multicolumn{1}{c}{-5.56}\\\
                         & \multicolumn{1}{c}{Large} & \multicolumn{1}{c}{37.01} & \multicolumn{1}{c}{-5.01}\\\
                         & \multicolumn{1}{c}{Near} & \multicolumn{1}{c}{36.18} & \multicolumn{1}{c}{-0.36}\\\
                         & \multicolumn{1}{c}{Far} & \multicolumn{1}{c}{$\>$ 40} & \multicolumn{1}{c}{-13.17}\\\hline
    \multirow{4}{*}{Bi}  & \multicolumn{1}{c}{Small} & \multicolumn{1}{c}{15.4} & \multicolumn{1}{c}{-10.92}\\\
                         & \multicolumn{1}{c}{Large} & \multicolumn{1}{c}{14.93} & \multicolumn{1}{c}{0.01}\\\
                         & \multicolumn{1}{c}{Near} & \multicolumn{1}{c}{16.15} & \multicolumn{1}{c}{-2.72}\\\
                         & \multicolumn{1}{c}{Far} & \multicolumn{1}{c}{12.97} & \multicolumn{1}{c}{-8.06}\\\hline
    \hline
  \end{tabular}
\end{table}

The collected data suggests drivers follow a common pattern for the studied case. At some distance
$D_{thresh}$ the drivers start veering away from the sub-lane which is closer to the hazard, they all
trace a path which maximizes their distance from the hazard and continue onwards or come back to 
the original lane depending upon traffic conditions. We found that for all the users in our population 
there was a statistically significant effect of size and position of hazard on driving behavior. 
Both our tests resulted in a p-value of less than 0.05. Interestingly, except one user 
everyone else collided with the occluded pedestrian at least once.

Figure \ref{fig:results} shows the generated constraints for cases from training set as well as OOTD
cases. Sub-figures (a) and (b) show the internal hazard-centric coordinate system
along with the resultant constraints in ego-car's coordinate system to illustrate the
transformation. Sub-figure (c) shows generated constraints for cases with multiple hazards
in sequence. The rightmost figure in sub-figure (c) is the hardest case that our algorithm
can handle.
Table \ref{table:insight} shows the distance of ego-car from
hazard at the maximum point of curvature of the preemptive trajectory as well as the 
calculated $D_{thresh}$ 
across the independent variables. 

First, we present a qualitative evaluation of the auto-generated constraints. 
Our criteria includes two factors: 1. Following rules of the road, 2. Being safety optimal 
in novel situations. For the former criteria, we would like to point out how the algorithm 
performs on bidirectional roads versus unidirectional roads. The constraints, completely
data-driven, are stricter for bidirectional scenario and the outer edge is more 
conservative as it tends to stick within its own lane allowance.

Now for the second, and more important factor, we would argue that the capability of the 
algorithm to handle multiple hazards in sequence attests to this. If one analyzes 
the figures a case can be made that the generated constraints ensure the trajectory is 
collision-free and realistic for a car to follow. 
To emphasize another interesting observation, one can see in the figures that the two 
extremes of the envelope are not actually symmetrical. While one is a tighter bound with 
less steering movement, the other is more curvy and weighted more towards ``evading'' the 
hazards. Such contrasting boundaries represent different stereotypes of driver profiles. 
The goal-oriented ones who take minimum deviations in evasive actions and the 
safety-oriented ones with extra steps of actions.

Next, we would like to briefly address the shortcoming of our algorithm here, namely its 
inability to handle hazards with higher degrees of overlap. We believe this is because humans
inherently treat single hazard and multi-hazard situations differently and the single hazard
demonstrations fail to capture this. The single hazard demonstrations tell us how far the 
ego-car should stay from the hazards, but when you add multiple such hazards this constraint
can turn the ego-car into a sitting duck, just like a mobile robot surrounded by multitudes 
in a crowd. Such situations require an inherent concept of aggression and ``goal-orientedness'' 
on the agent's part, which is very different than what our demonstrations show.

Moving to the second part of our study, our statistical analysis suggests a validation of our 
initial hypothesis. It is also interesting to
note here that as per \citet{berndt1994using} such studies in simulation with results consistent across
variables and users, are especially suited for evaluating hazard perception of the users and by an
extension learning from the good measures.
Moreover, hazard perception ability in humans has been found to be directly linked to
risk of accident by the driver \cite{mckenna1999hazard}.

A striking observation that can be gleaned from table \ref{table:insight} is the variation in
point of maximum curvature for positional cases under unidirectional road condition
and for size under bidirectional condition.
Under the bidirectional condition, the smaller hazard still allows the driver to have a largely
unobstructed view of the road, see figure \ref{fig:hazards}.
In accordance with our hypothesis of preemptive behavior, the user here steers away well ahead
in advance and is able to maintain a forward trajectory without colliding with either the hazard
or the oncoming traffic while maintaining view of the road itself.
However the larger hazard ends up blocking most of the user's access and view of the lane.
In order to abide the road rules and prevent collision, the user here traces a larger curve 
trajectory but only when right next to the hazard.
This is because this larger trajectory requires creeping into the next lane first before heading
back on one's own.

This is an interesting case study for modeling corner-cases in AVs since our observations from
real human users suggest that sometimes for the safety of ego as well as other agents in
the environment, the rules of the road need to be made fluid.
There needs to be a model which can inform the trade-off between such preemptive behaviors
versus strict adherence to traffic rules depending upon scene-understanding.
Under the current bottleneck of imperfect perception, we advocate the use of scientific
exploration and analysis of corner-cases in urban driving scenarios to push the
needle a little further in terms of increased autonomy with safety guarantees.

By the way of this paper, this is also our attempt in advocating for the use of LfD or imitation
based techniques to encompass the complex decision process of a human driver.
AVs are specially well suited for this problem, since recording ``good'', safe trajectories on
the platform itself is much easier as compared to traditional robotics arms or even mobile
platform operated via game controllers.
This means the embodiment mapping \cite{argall2009survey} for the data is effectively the identity,
thereby reducing the scope of our efforts to solve only the record mapping problem, i.e. states observed
by machine to be similar to what the human observes.

\bibliography{bibliography}

\begin{thebibliography}{16}
\providecommand{\natexlab}[1]{#1}
\providecommand{\url}[1]{\texttt{#1}}
\expandafter\ifx\csname urlstyle\endcsname\relax
  \providecommand{\doi}[1]{doi: #1}\else
  \providecommand{\doi}{doi: \begingroup \urlstyle{rm}\Url}\fi

\bibitem[Akgun et~al.(2012)Akgun, Cakmak, Jiang, and Thomaz]{akgun2012keyframe}
Baris Akgun, Maya Cakmak, Karl Jiang, and Andrea~L Thomaz.
\newblock Keyframe-based learning from demonstration.
\newblock \emph{International Journal of Social Robotics}, 4\penalty0
  (4):\penalty0 343--355, 2012.

\bibitem[Argall et~al.(2009)Argall, Chernova, Veloso, and
  Browning]{argall2009survey}
Brenna~D Argall, Sonia Chernova, Manuela Veloso, and Brett Browning.
\newblock A survey of robot learning from demonstration.
\newblock \emph{Robotics and autonomous systems}, 57\penalty0 (5):\penalty0
  469--483, 2009.

\bibitem[Atkeson and Schaal(1997)]{atkeson1997robot}
Christopher~G Atkeson and Stefan Schaal.
\newblock Robot learning from demonstration.
\newblock 97:\penalty0 12--20, 1997.

\bibitem[Berndt and Clifford(1994)]{berndt1994using}
Donald~J Berndt and James Clifford.
\newblock Using dynamic time warping to find patterns in time series.
\newblock 10\penalty0 (16):\penalty0 359--370, 1994.

\bibitem[Bojarski et~al.(2016)Bojarski, Del~Testa, Dworakowski, Firner, Flepp,
  Goyal, Jackel, Monfort, Muller, Zhang, et~al.]{bojarski2016end}
Mariusz Bojarski, Davide Del~Testa, Daniel Dworakowski, Bernhard Firner, Beat
  Flepp, Prasoon Goyal, Lawrence~D Jackel, Mathew Monfort, Urs Muller, Jiakai
  Zhang, et~al.
\newblock End to end learning for self-driving cars.
\newblock \emph{arXiv preprint arXiv:1604.07316}, 2016.

\bibitem[Brodlie and Butt(1991)]{brodlie1991preserving}
KW~Brodlie and Sohail Butt.
\newblock Preserving convexity using piecewise cubic interpolation.
\newblock \emph{Computers \& Graphics}, 15\penalty0 (1):\penalty0 15--23, 1991.

\bibitem[Chen et~al.(2015)Chen, Seff, Kornhauser, and
  Xiao]{chen2015deepdriving}
Chenyi Chen, Ari Seff, Alain Kornhauser, and Jianxiong Xiao.
\newblock Deepdriving: Learning affordance for direct perception in autonomous
  driving.
\newblock pages 2722--2730, 2015.

\bibitem[Fritsch and Carlson(1980)]{fritsch1980monotone}
Frederick~N Fritsch and Ralph~E Carlson.
\newblock Monotone piecewise cubic interpolation.
\newblock \emph{SIAM Journal on Numerical Analysis}, 17\penalty0 (2):\penalty0
  238--246, 1980.

\bibitem[Isele et~al.(2017)Isele, Cosgun, Subramanian, and
  Fujimura]{isele2017navigating}
David Isele, Akansel Cosgun, Kaushik Subramanian, and Kikuo Fujimura.
\newblock Navigating intersections with autonomous vehicles using deep
  reinforcement learning.
\newblock \emph{arXiv preprint arXiv:1705.01196}, 2017.

\bibitem[McKenna and Horswill(1999)]{mckenna1999hazard}
Frank~P McKenna and Mark~S Horswill.
\newblock Hazard perception and its relevance for driver licensing.
\newblock \emph{IATSS research}, 23\penalty0 (HS-042 879), 1999.

\bibitem[Mitsch et~al.(2013)Mitsch, Ghorbal, and Platzer]{mitsch2013provably}
Stefan Mitsch, Khalil Ghorbal, and Andr{\'e} Platzer.
\newblock On provably safe obstacle avoidance for autonomous robotic ground
  vehicles.
\newblock 2013.

\bibitem[Ross et~al.(2011)Ross, Gordon, and Bagnell]{ross2011reduction}
St{\'e}phane Ross, Geoffrey~J Gordon, and Drew Bagnell.
\newblock A reduction of imitation learning and structured prediction to
  no-regret online learning.
\newblock pages 627--635, 2011.

\bibitem[Sadigh et~al.(2016)Sadigh, Sastry, Seshia, and
  Dragan]{sadigh2016planning}
Dorsa Sadigh, Shankar Sastry, Sanjit~A Seshia, and Anca~D Dragan.
\newblock Planning for autonomous cars that leverage effects on human actions.
\newblock 2016.

\bibitem[Wei et~al.(2013)Wei, Snider, Kim, Dolan, Rajkumar, and
  Litkouhi]{wei2013towards}
Junqing Wei, Jarrod~M Snider, Junsung Kim, John~M Dolan, Raj Rajkumar, and
  Bakhtiar Litkouhi.
\newblock Towards a viable autonomous driving research platform.
\newblock pages 763--770, 2013.

\bibitem[Zhang et~al.(2016)Zhang, Kahn, Levine, and Abbeel]{zhang2016learning}
Tianhao Zhang, Gregory Kahn, Sergey Levine, and Pieter Abbeel.
\newblock Learning deep control policies for autonomous aerial vehicles with
  mpc-guided policy search.
\newblock pages 528--535, 2016.

\bibitem[Ziebart et~al.(2008)Ziebart, Maas, Bagnell, and
  Dey]{ziebart2008maximum}
Brian~D Ziebart, Andrew~L Maas, J~Andrew Bagnell, and Anind~K Dey.
\newblock Maximum entropy inverse reinforcement learning.
\newblock 8:\penalty0 1433--1438, 2008.

\end{thebibliography}

\end{document}